\documentclass[conference]{IEEEtran}
\IEEEoverridecommandlockouts
\usepackage{cite}
\usepackage{amsmath,amssymb,amsfonts}
\usepackage[T5]{fontenc}
\usepackage{algorithmic}
\usepackage{graphicx}
\usepackage{textcomp}
\usepackage{xcolor}
\usepackage{multirow}
\usepackage{balance}
\usepackage{authblk}
\usepackage{array}
\usepackage{float}
\usepackage{stfloats}
\usepackage[pagebackref=true,breaklinks=true,letterpaper=true,colorlinks,bookmarks=false]{hyperref}
\usepackage[utf8]{inputenc}
\usepackage{tabularx}
\usepackage{adjustbox}

\def\BibTeX{{\rm B\kern-.05em{\sc i\kern-.025em b}\kern-.08em
    T\kern-.1667em\lower.7ex\hbox{E}\kern-.125emX}}
    
\begin{document}
% \title{How To Create Your Own Assistant: Efficient Method Based On Large Language Model For Vietnamese\\
% }
\title{Efficient Finetuning Large Language Models For Vietnamese Chatbot}
\author[1,2]{Vu-Thuan Doan}
\author[1,2]{Quoc-Truong Truong}
\author[1,2]{Duc-Vu Nguyen}
\author[1,2,*]{Vinh-Tiep Nguyen}
\author[1,2]{Thuy-Ngan Nguyen Luu}
\affil[1]{University of Information Technology, Ho Chi Minh City, Vietnam}
\affil[2]{Vietnam National University, Ho Chi Minh City, Vietnam}
\affil[ ]{thuandv.14@grad.uit.edu.vn, \{ truongtq, vund, tiepnv,ngannlt\}@uit.edu.vn}
\affil[*]{Corresponding author}
\maketitle

\begin{abstract}
% Cau dau thuong noi ve bai toan / dong luc của hướng tiếp cận bài báo này.
% Bài toán chính mà mình quan tâm giải quyết là gì?
% Vấn đề của bài toán này đang gặp phải là gì
% Phương pháp tiếp cận của bài báo này là gì?
% Kết quả thực nghiệm của bài báo này có gì thú vị.

Large language models (LLMs), such as GPT-4, PaLM, and LLaMa, have been shown to achieve remarkable performance across a variety of natural language tasks. Recent advancements in instruction tuning bring LLMs with ability in following user's instructions and producing human-like responses. 
% However, the expensive training and deployment of LLMs present challenges to open academic research. 
However, the high costs associated with training and implementing LLMs pose challenges to academic research.
% Furthermore, the availability of fine-tuned LLMs for the Vietnamese language is limited.
Furthermore, the availability of pretrained LLMs and instruction-tune datasets for Vietnamese language is limited.
% To address these issues, we 
% % collect: 
% leverage large-scale datasets from English
% and translate instruction-following datasets for generic and medical domains from open-source projects: Alpaca, GPT4All, and ChatDoctor. 
To tackle these concerns, we leverage large-scale instruction-following datasets from open-source projects, namely Alpaca, GPT4All, and Chat-Doctor, which cover general domain and specific medical domain.
To the best of our knowledge, these are the first instructional dataset for Vietnamese.
Subsequently, we utilize parameter-efficient tuning through Low-Rank Adaptation (LoRA) on two open LLMs: Bloomz (Multilingual) and GPTJ-6B (Vietnamese), resulting four models: Bloomz-Chat, Bloomz-Doctor, GPTJ-Chat, GPTJ-Doctor.
% To the best of our knowledge, this is the first instructional datasets and pre-trained models for Vietnamese. 
% Dua quy mo cua dataset vao
% Finally, we evaluate our method by utilizing GPT-3.5 as a scoring method for each sample, which is more cost-efficient than human evaluation. 
% Finally, we evaluate our approach by employing GPT-3.5 as a cost-effective scoring method for each sample based on the helpfulness, relevance, accuracy, level of details of their responses, eliminating the need for human evaluation. 
% The result shows that even with a very low-cost setup, our method can archive the score of 77 generated by our scoring method.
Finally, we assess the effectiveness of our methodology on a per-sample basis, taking into consideration the helpfulness, relevance, accuracy, level of detail in their responses. This evaluation process entails the utilization of GPT-4 as an automated scoring mechanism.
% , eliminating the necessity for labor-intensive human assessment.
%With only 6 hours of finetuning  using 4 RTX-4090 GPU, 
% Neu muon dua so vao thi dua them so cua nhung thang khac
% Our method can achieve the score of 77\% for Bloomz-Chat and 87\% for Bloomz-Doctor.
% Even with a very low-cost setup, our technique can score 77\% for Bloomz-Chat and 87\% for Bloomz-Doctor.
% Despite utilizing an low-cost setup, our method demonstrate an improvement of 30\% with Bloomz and 20\% with GPTJ-6B.achieves a scoring of 73.55\% for Bloomz-Chat and 74.92\% for Bloomz-Doctor.
Despite utilizing a low-cost setup, our method demonstrates about 20-30\% improvement over the original models in our evaluation tasks.
% chua thay noi ten metric, va co so sanh voi hien trang truoc khi finetune
% CAN NHAN MANH: to the best of our knowledge, this is the first open Vietnamese dataset and pre-trained model --> day la word dau tien open finetuned model cho tieng Viet
\end{abstract}


\section{Introduction}
% Language modeling has long been an important research area since Shannon (1951) estimated the information in
% language with next word prediction. Modeling began with n-gram based approaches (Kneser & Ney, 1995) but rapidly
% advanced with LSTMs (Hochreiter & Schmidhuber, 1997; Graves, 2014). Later work showed that language modelling
% also led to language understanding (Dai & Le, 2015). With increased scale and the Transformer architecture (Vaswani
% et al., 2017), large language models (LLMs) have shown strong performance in language understanding and generation
% capabilities over the last few years, leading to breakthrough performance in reasoning, math, science, and language tasks
% (Howard & Ruder, 2018; Brown et al., 2020; Du et al., 2022; Chowdhery et al., 2022; Rae et al., 2021; Lewkowycz
% et al., 2022; Tay et al., 2023; OpenAI, 2023b). Key factors in these advances have been scaling up model size (Brown
% et al., 2020; Rae et al., 2021) and the amount of data (Hoffmann et al., 2022). To date, most LLMs follow a standard
% recipe of mostly monolingual corpora with a language modeling objective

In recent years, large language models (LLMs) have garnered significant attention thanks to their remarkable success in numerous natural language processing (NLP) tasks. 
Large language models are based on a deep learning architecture known as a Transformer. The Transformer model revolutionized natural language processing tasks by effectively capturing long-range dependencies and relationships within text. 
LLMs are trained on large volumes of text data to predict the subsequent tokens, enabling them to generate coherent and fluent text in response to various inputs. 
However, without few-shot exemplars, it is harder for models to perform well on prompts that are not similar to the format of the pretraining data. These models also struggle to follow instructions or goals specified by users, which limits their usefulness and applicability in real-world scenarios.

The NLP community has recently witnessed many endeavors to train large language models to follow instructions better and be more helpful. 
Large “instruction-tuned” language models
(finetuned to respond to instructions) have
demonstrated a remarkable ability to generalize zero-shot to new tasks.
Initial attempts \cite{sanh2021multitask} \cite{wang2022super} \cite{wei2021finetuned} \cite{chung2022scaling} to train instruction-following language models are based on a collection of various NLP tasks, with a set of human-annotated instructions accompanying each task. These developments are powered by two key components: large pre-trained language models and human-written instruction data. However, this process is costly and often suffers limited diversity in NLP tasks. Then, Self-Instruct \cite{wang2022self}, a framework for improving the instruction-following capabilities of pre-trained language models by bootstrapping off its own generations. 
% This method has made significant improvements for LLMs in following instructions in natural language and can generate professional and contextual responses in a conversational way. 
This method has addressed such challenges necessitates the creation of a large-scale, public dataset covering a broad range of tasks.

Despite the great success of LLMs, SOTA models like GPT-4, and PaLM \cite{chowdhery2022palm}, are often only accessible through restricted APIs, creating barriers for new research endeavors. In the recent literature, there has been a growing interest in leveraging open-source LLMs and adapting them towards specific applications or domains. Popular models are Bloom \cite{muennighoff2022crosslingual} from Bigscience, GPT-J from EleutherAI, and LLaMa \cite{touvron2023llama} from Meta. 
% However, in Vietnamese, we don't have any language model which has trained on a very large corpus of text to have LLM's characteristic: a few short generations with in-context learning.
Nonetheless, within the Vietnamese context, there is a lack of language models that have undergone extensive training on vast corpora, resulting in the absence of the distinctive features possessed by LLMs: a few short generations with in-context learning.

Another challenge when working with LLMs is the cost of fine-tuning and deployment. To fine-tune large language models in a low-resource setting, we utilize a parameter-efficient
tuning approach that effectively leverages the limited computational resources available. By applying LoRA(Low-Rank Adaptation) technique, this enables the adaptation of state-of-the-art language models to resource-constrained scenarios while maintaining high performance and adaptability.

\begin{figure*}[!t]
% \centerline{\includegraphics[width=\textwidth,height=\textheight,keepaspectratio]{fig/fig1.png}}
\centerline{\includegraphics[width=1\linewidth]{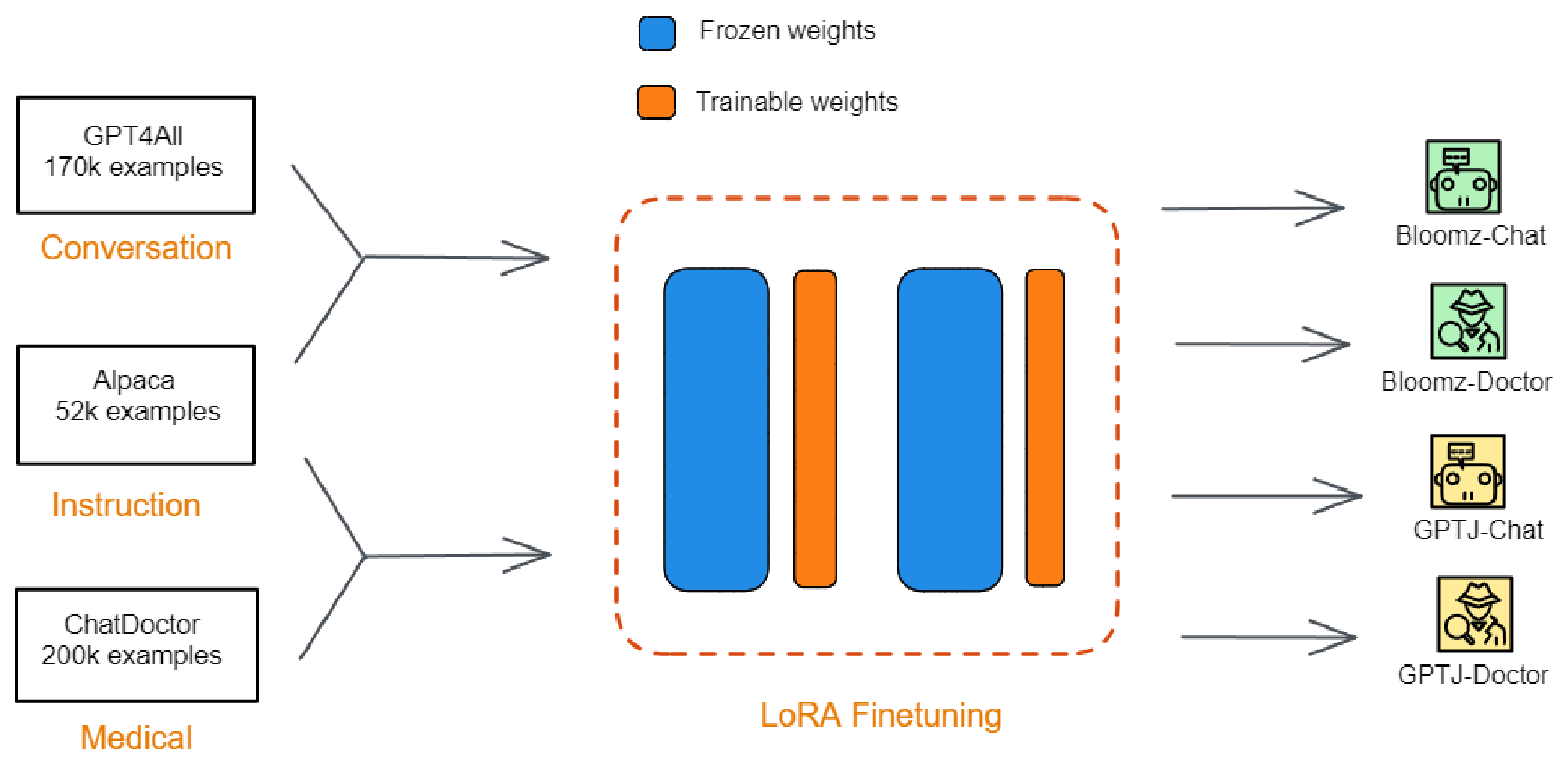}}
\caption{The overview of our approach using Instruction Tuning with LoRA.}
\label{overal-arch}
\end{figure*}

In this paper, we propose a simple method based on efficient instruct-tuning with our Vietnamese instruction dataset to build two chatbot models for generic and specific domains, e.g. medical. The overview of our approach is shown in {Figure \ref{overal-arch}}.
%We also realease our experimental code and model weights in github link.
Our main contributions in this paper are as follows:
\begin{itemize}
    \item We propose new instruction-following datasets for generic and medical domains by collecting and translating from other public sources.
    
    \item We adopt the Low-Rank Adaptation (LoRA) approach for the efficient training and deployment of LLMs. This makes the training and deployment cost affordable for academic research.

    \item We evaluate the performance of those approaches on a collection of natural language tasks, demonstrating the ability in the context of Vietnamese for general and medical domains.
\end{itemize}

% The contribution of our work is that we show ViT is capable of face recognition task and give comparable results with CNN model and even better results if there are not enough data in training phase. We use CosFace not only as the loss function but also as the face classification and it gives better results in the finetuning process when compared with other methods such as L2 distance and SVM (Support Vector Machine). We also show that if there are not enough data In the finetuning process (in this case we use 1 image per person to train the models), the results are worse than those from pre-trained models with MS-Celeb-1M. But if we use 2 images per person to finetune, the results are better.

\section{Related Work}
% Một số hướng tiếp cận chính của nhận diện gương mặt gần đây
% Đề cập đến Hướng tiếp cận ViT cho bài toán phân loại văn bản. Sau đó là ViT cho bài toán nhận diện gươn  mặt.
% Bài báo này có gì khác so với các hướng tiếp cận trước đây. Làm rõ sự khác biệt là được.
% Đặt lại tiêu đề nghe hấp dẫn hơn từ Using
% Ví dụ: Baby learning with Vision Transformer for Face Recognition

\textbf{Large Language Models:} have demonstrated impressive capabilities across various domains, including natural language understanding, text generation, dialogue systems, content summarization, and more. 
% They have found applications in areas like content creation, virtual assistants, customer support chatbots, language translation, and even creative writing.
Recent advancements in large language models (LLMs) have demonstrated their superiority over previous-generation paradigms, such as pretraining and fine-tuning. The significant increase in model scale has led to qualitative changes in LLMs, commonly referred to as emergent abilities. These include in-context learning for zero-shot tasks and chains of thought that enhance the model’s performance on complex tasks. 
OpenAI’s development of ChatGPT and GPT-4 \cite{openai2023gpt4} has revolutionized the perception of LLMs. Although these models exhibit remarkable performance, OpenAI has not disclosed details regarding their training strategies or weight parameters. However, there are several open-source LLMs alternatives for GPT-4: Bloom\cite{scao2022bloom}, Bloomz \cite{muennighoff2022crosslingual}, GPT-J, and LLaMa\cite{touvron2023llama} with sizes ranging from 7B to 65 billion parameters. 

\textbf{Instruction Tuning:} The subfield of language models concentrates on the
instruction-following capabilities is crucial for generating responses based on natural language commands. Instruction-following methods enhance pre-trained models by fine-tuning them using high-quality input-output tuples of task instructions and ground truth outputs. This finetuning helps the model better understand user intentions and follow instructions more accurately. Instruction-following methods have been extensively researched in language models \cite{sanh2021multitask} \cite{wang2022super} \cite{wei2021finetuned} \cite{chung2022scaling} and multi-modality domains \cite{muennighoff2022crosslingual}. Among those methods, FLAN \cite{wei2021finetuned} introduces an instruction-tuning method that outperforms non-tuned LLMs in unseen tasks. PromptSource \cite{bach2022promptsource} provides a development environment and repository that offers a web-based GUI for creating and managing natural language prompts for zero-shot or gradient-based few-shot learning. SUP-NATINST \cite{wang2022super} establishes a large benchmark of 1,616 diverse NLP tasks and uses multi-task training on the T5 model and demonstrates strong generalization capabilities on unseen tasks. InstructGPT \cite{ouyang2022training} demonstrates significant performance improvements and may be integrated into closed-source models like GPT-3.5 and GPT-4 \cite{openai2023gpt4}. The open-source \href{https://crfm.stanford.edu/2023/03/13/alpaca.html}{Stanford Alpaca} approach fine-tunes all parameters of LLMs in an end-to-end manner.

% \begin{figure*}[!h]
% \centerline{\includegraphics[width=8cm,keepaspectratio]{fig/param-effect.jpg}}
% \caption{Parameter-Efficient Transfer Learning Methods.}
% \label{overal-arch}
% \end{figure*}
\textbf{Parameter-Efficient Fine-Tuning:} Parameter-Efficient Fine-Tuning (PEFT) \cite{chen2023parameter} 
% methods facilitate efficient adaptation of LLMs without the need to update all model parameters, thereby reducing the cost and improving the efficiency of fine-tuning large models. 
aims to optimize the fine-tuning process by efficiently utilizing the available computing resources and reducing the number of parameters that need to be updated. This approach becomes particularly relevant when working with limited labeled data for a specific task. This approach not only saves computational time and resources but also enables the deployment of large language models  more accessible and practical for a wide range of applications.
Various PEFT techniques include Prefix Tuning \cite{li2021prefix}, Low-Rank adaptation (LoRA) \cite{hu2021lora}, and the insertion of adapter layers in pre-trained large language models. Prefix Tuning \cite{li2021prefix} appends a collection of prefixes to autoregressive language models, or alternatively, incorporates prefixes for both encoder and decoder components, similar methods proposed in \cite{lester2021power}. LoRA \cite{hu2021lora} introduces trainable rank decomposition matrices into each layer. Adapters involve inserting lightweight modules into each layer of pre-trained models, which only updates the adapters and has been extended across numerous domains.

\section{Method}

In this section, we will describe our method to collect datasets, choose pretrained language model and how we apply parameter-efficient fine-tuning.

\subsection{Data Collection}
As creating an instruction tuning dataset with many tasks from scratch would be resource-intensive, we transform existing datasets from the research community into an instructional format and translate them into Vietnamese
\\
Instruction dataset example:
\\
\noindent\fbox{%
    \parbox{\linewidth}{%
        \emph{ 
        \{"prompt": "đưa ra 3 mẹo giữ sức khỏe", "response": "1. Ăn một chế độ ăn cân bằng và đảm bảo bao gồm nhiều trái ..."\}\\
        \{"prompt": "Ba màu chính là gì?", "response": "Ba màu chính là màu đỏ, xanh và vàng."\}\\
        \{"prompt": "Mô tả cấu trúc của một nguyên tử.", "response": "Một nguyên tử được tạo thành từ một hạt nhân, chứa..."\}
        }
    }%
}%
\\
\subsubsection{Instruction-Following Dataset}
First, we collected instruction-following examples from the Standford Alpaca project. This project adopt the Self-Instruct \cite{wang2022self} technique and generated 52k demonstrations from the open-source LLaMa-7B model instead of GPT3.5 in the original Self-Instruct \cite{wang2022self} paper.

\begin{figure}[h]
    \centering
    \includegraphics[width=0.8\linewidth, keepaspectratio]{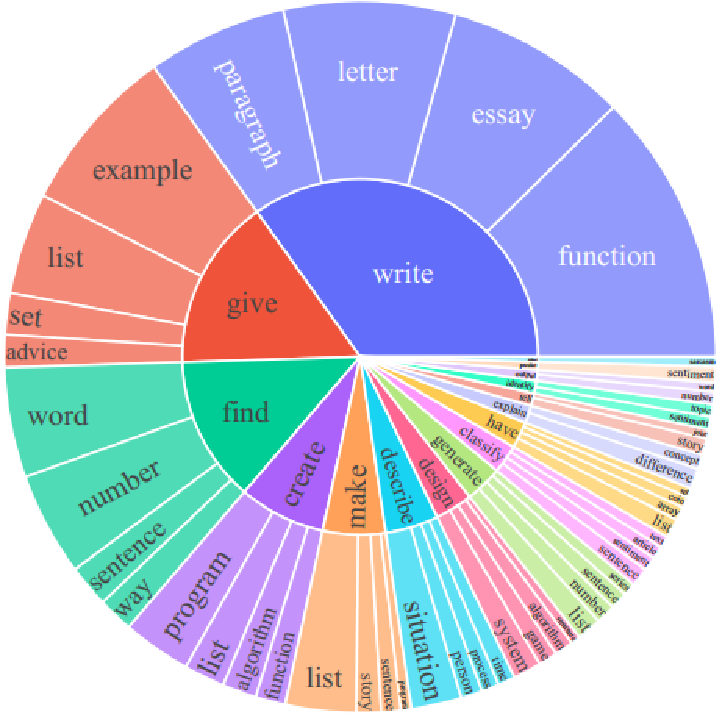}
    \caption{Apaca Task Distribution \cite{wang2022self} } 
    \label{task-distribution}
\end{figure}

 To increase more instruction data, we also collect data from the GPT4All project.

\subsubsection{Medical Dataset} 
% We followed ChatDoctor \cite{yunxiang2023chatdoctor} method to build medical dataset. It starts with a database containing about 700 diseases with their relative symptoms, medical tests, and recommended medications.
% Then it uses ChatGPT to generate 5k conversations between patients and physicians from that seed database.
% Moreover, we collect 195k real conversations between patients and doctors from \href{HealthCareMagic.com}{Health Care Magic} and combine them into instructional dataset for medical domain. Finally, we translate all data to Vietnamese using Google Translate API. 
For medical instruction dataset, we collect from ChatDoctor \cite{yunxiang2023chatdoctor}. Their method starts with a database containing about 700 diseases with their relative symptoms, medical tests, and recommended medications.
Then it uses ChatGPT to generate 5k conversations between patients and physicians from that seed database.
We also collect 195k real conversations between patients and doctors from \href{HealthCareMagic.com}{Health Care Magic} and combine them into instructional dataset for medical domain. Finally, we translate all data to Vietnamese using Google Translate API. 
\\
\\
Medical dataset example:
\\
\noindent\fbox{%
    \parbox{\linewidth}{%
        \emph{ 
        \{"prompt": "Thưa bác sĩ, tôi bị nôn, buồn nôn và đau bụng. Gần đây tôi phát hiện ra rằng tôi có môn vị hẹp. Tôi nên dùng thuốc gì cho nó?", \\
        "response": "Phương pháp điều trị hẹp môn vị phổ biến nhất là một thủ thuật phẫu thuật được gọi là phẫu thuật cắt bỏ môn vị. Điều này liên quan đến
cắt và niêm phong cơ xung quanh lỗ mở của
môn vị, mở rộng nó và cho phép thức ăn đi qua
dễ dàng hơn. Tuy nhiên cũng có một số thuốc
có thể giúp giảm các triệu chứng như metoclopramide hoặc cisapride. Bạn thích cái nào hơn?"\}
        }
    }%
}%

\begin{table}[h]
\caption{\label{tab:dataset_statistic} Dataset statistic}
\begin{center}
\def\arraystretch{1.3}
\begin{tabular}{c c c c}
\hline
\textbf{Dataset} & \textbf{Samples} & \textbf{Avg. tokens/sample} & \textbf{Translated Tokens} \\
\hline
Alpaca  & 52k  & 350  & 18M\\
\hline    
GPT4All  & 150k  & 365  & 61M\\
\hline 
ChatDoctor  & 200k  & 405 &  87M\\
\hline 
\end{tabular}
\end{center}
\end{table}

\subsection{Baseline Models}
% There are a lot of good open-source LLMs: BERT, T5, GPT-J, Bloom, and LLaMA. However, only a few of them have trained with enough Vietnamese text corpus and have trained as a casual language model (decoder only) which is suitable for text generation tasks. Then we picked Bloomz-mt-7B \cite{muennighoff2022crosslingual} as our baseline model. 
We use publicly available pretrained Bloomz-mt-7B \cite{muennighoff2022crosslingual} and \href{https://huggingface.co/VietAI/gpt-j-6B-vietnamese-news}{GPTJ-6B} from VietAI as our baseline models. Bloomz is a small variant of original BLOOM model, a collaborative project of more than 1,000 scientists and the amazing Hugging Face team. 
% It is openly available for everybody. 
The BLOOM model is an open-access multilingual language model that contains 176B parameters and was trained for 3.5 months on 384 A100–80GB GPUs.It's small variant Bloomz is a multi-language model which has trained on the ROOTS corpus and finetuned with the xP3 dataset. This dataset has 59 Languages (46 natural and 13 programming languages including 3\% Vietnamese). GPTJ-6B from VietAI is the one of biggest GPT models trained only on Vietnamese dataset.

% \begin{figure}[h]
%     \centering
%     \includegraphics[width=1.0\linewidth, keepaspectratio]{fig/bloom_network_1.png}
%     \caption{BLOOM architecture}
%     \label{bloom-arch}
% \end{figure}

\subsection{Parameter-Efficient Tuning}
\begin{figure}[h]
    \centering
    \includegraphics[width=1.0\linewidth, keepaspectratio]{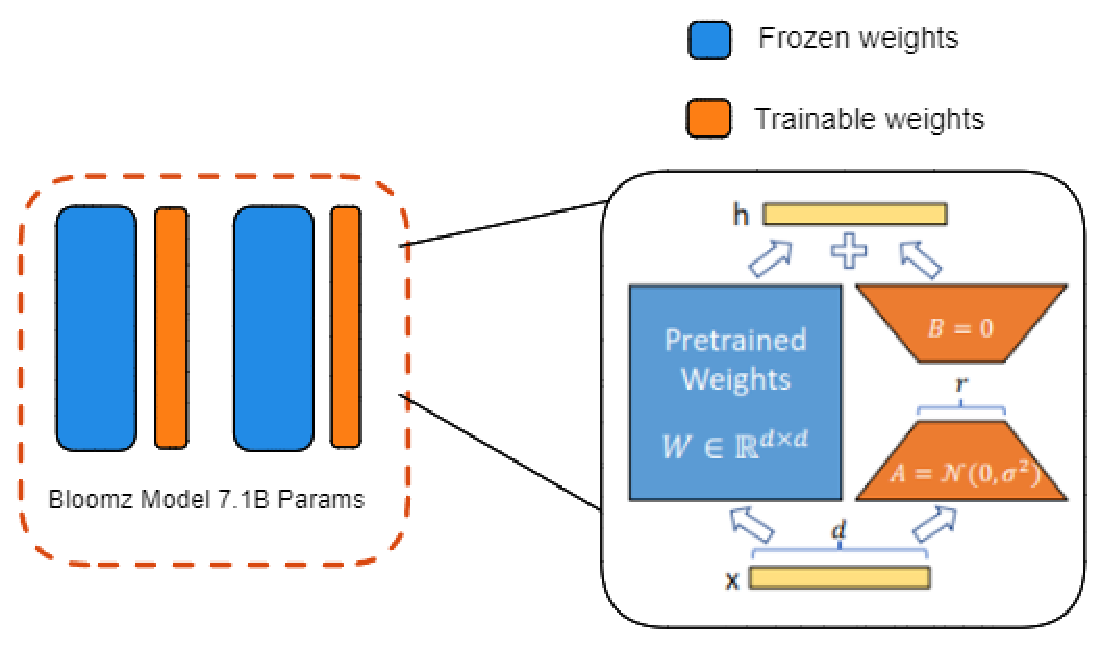}
    \caption{The overview of our approach using Instruction Tuning with LoRA.}
    \label{lora-arch}
\end{figure}

Standard fine-tuning often requires vast amounts of computational resources, as well as high-quality and extensive datasets. However, given the limited availability of high-quality multi-turn chat corpora, it is crucial to adopt methods that are more efficient in terms of computational cost and data requirements. Parameter-efficient tuning methods \cite{chen2023parameter} help achieve this goal by making better use of the available data and minimizing the need for extensive resource allocation. More details in {Figure \ref{lora-arch}}.

Specifically, we use Low-Rank Adaption (LoRA, Hu et al.) to fine-tune the base LLMs. For a linear layer \(h = W_0x\), the forward pass is modified to be:

\begin{equation}
    h = W_0x + BAx
\end{equation}

where \(W_0 \in R^{d \times k}, B \in R^{d \times r}, A \in R^{r\times k}\) with the rank \(r \ll min(d; k) \)
\\
\section{Experiments}
\subsection{Training Dataset}

We construct three separate datasets for our models. For generic chatbot, we combine data from Alpaca and GPT4All into one dataset. For medical chatbot, we combine 52k data samples from Alpaca with 200k conversations from ChatDoctor as training dataset. We format the dataset as json file as GPT-3 style that includes two fields of instruction and response for each sample. The prompt template is as simple as follows.

\emph{
\\
Prompt template for Bloomz-Chat and GPTJ-Chat:
\\
}

\noindent\fbox{%
    \parbox{\linewidth}{%
        \emph{
        Hãy viết một phản hồi thích hợp cho chỉ dẫn dưới đây.
        \\
        \#\#\# Instruction:\\
        \{instruction\}
        \\
        \#\#\# Response:
        }
    }%
}%

\emph{
\\
\\
Prompt template for Bloomz-Doctor and GPTJ-Doctor:
\\
}

\noindent\fbox{%
    \parbox{\linewidth}{%
        \emph{
        Nếu bạn là bác sĩ, vui lòng trả lời các câu hỏi y tế dựa trên mô tả của bệnh nhân.
        \\
        \#\#\# Instruction:\\
        \{instruction\}
        \\
        \#\#\# Response:
        }
    }%
}%
\\
\\
\subsection{Training Details}

We set up the environment with PyTorch and Huggingface Transformers package. Parameter-efficient fine-tuning was applied on our models based on the codebase of \href{https://github.com/tloen/alpaca-lora}{Alpaca-Lora}. The maximum length of the input sequence to 512 and the rank k in LoRA to 8. The base model checkpoints were initialized with the 8-bit integer format (int8) parameters released by Touvron et al., which remain fixed during training, thus reducing GPU memory consumption and improving training speed. We use the Adam optimizer to update LoRA parameters with a total batch size of 128 and learning rates of 3e-4. The trainable LoRA parameters are about 4.2M parameters and fine-tuned for 2 epochs on 4 RTX-4090-24GB GPU. The training time is listed in {Table \ref{tab:training_details}}.

\begin{table*}
\caption{\label{tab:training_details} Training Details}
\begin{center}
\def\arraystretch{1.5}
\begin{tabular}{c c c c c}
\hline
\textbf{Model} & \textbf{Original Param.} & \textbf{Trainable Param.} & \textbf{Training Time} %& \textbf{Training Cost} 
& \textbf{Dataset} \\
\hline
Bloomz-Chat  & 7.1B  & 4.2M & 6h %& \$12 
& Alpaca + GPT4All \\
\hline
Bloomz-Doctor  & 7.1B  & 4.2M & 6h30m %& \$13 
& Alpaca + ChatDoctor \\
\hline
GPTJ-Chat  & 6B  & 3.6M & 5h %& \$12 
& Alpaca + GPT4All \\
\hline
GPTJ-Doctor  & 6B  & 3.6M & 5h30m %& \$13 
& Alpaca + ChatDoctor \\
\hline
\end{tabular}
\end{center}
\end{table*}

\subsection{Inference Details}

During the inference phase, we also use an inference prompt as we did for the training stage to improve conversational capabilities. For the decoding strategy, we use these settings with the details are as follows.

\begin{itemize}
    \item Max new tokens: We limit max new tokens to 256 tokens to ensure that the outputs remain focused and relevant to the input prompt.
    
    \item Temperature: We set the temperature to 0.5, which controls the randomness of the sampling process. Lower values make the model generate more focused and deterministic outputs, while higher values increase diversity at the cost of coherence.

    \item Top-k sampling: We use Top-k sampling with k = 20, meaning that the model selects its next token from the top 20 most probable tokens at each step, adding an element of randomness and diversity to the generated text.

    \item Repetition penalty: To discourage the model from generating repetitive text, we apply a repetition penalty with a factor of 1.2, penalizing tokens that have already been selected.
\end{itemize}

\section{Evaluation}
Evaluating the performance of text generation tasks can be challenging due to the significant variety in their form, unlike natural language understanding tasks (such as text classification and extractive machine reading comprehension). Following previous works (\cite{yunxiang2023chatdoctor}, \href{https://lmsys.org/blog/2023-03-30-vicuna/}{Vicuna})  that utilizes GPT-4 as a scoring method, we also adopt GPT-4 to provide an overall score (on a 100-point scale) for each sample, which is more efficient than human evaluation. 
% However, GPT-4 may not always provide accurate scores, so we perform manual checks on its ratings and adjust them if necessary. The manual checks ensure that the scores are consistent and reflect the true performance of the models being evaluated. 

The evaluation system judge each response from our models by four aspects:
\begin{itemize}
    \item Relevance: Assessing the model’s ability to correctly interpret the semantic meaning of the context and questions.
\item Helpfulness: Assessing the model’s ability to provide useful information.
\item Accuracy: Evaluating whether the model can perform correctly in the corresponding for a given instruction.
\item Level of details: Whether the model can accurately use various and detailed knowledge for problem.
\end{itemize}
We use the following prompt template for scoring the outputs of the systems:\\

% \begin{figure}[h]
%     \centering
%     \includegraphics[width=1.0\linewidth, keepaspectratio]{fig/eval_prompt.png}
%     \caption{Evaluation Prompt Template}
%     \label{lora-arch}
% \end{figure}

\noindent\fbox{%
    \parbox{\linewidth}{%
        \emph{
        System Prompt: You are a helpful and precise assistant for checking the quality of the answer \\
        Prompt: \\
        \#\#\#Question\\
        \{question\}\\
        \#\#\#The Start of Assistant 1's Answer\\
        \{answer\_1\}\\
        \#\#\#The End of Assistant 1's Answer\\
        \#\#\#The Start of Assistant 2's Answer\\
        \{answer\_2\}\\
        \#\#\#The End of Assistant 2's Answer\\
        We would like to request your feedback on the performance of two AI assistants in response to the user question displayed above. \\
        Please rate the helpfulness, relevance, accuracy, level of details of their responses. Each assistant receives an overall score on a scale of 1 to 10, where a higher score indicates better overall performance... \\
        % Please first output a single line containing only two values indicating the scores for Assistant 1 and 2, respectively. The two scores are separated by a space. In the subsequent line, please provide a comprehensive explanation of your evaluation, avoiding any potential bias and ensuring that the order in which the responses were presented does not affect your judgment.
        }
    }%
}%
\\
\\
Our evaluation set is designed to provide a comprehensive assessment of our models across a wide range of natural language understanding and generation tasks. The set comprises 80 samples, covering 9 distinct categories, including Question Answering, Reasoning, Literature, Entertainment, Math, and Coding. The overall score for a specific task is calculated by summing the scores for all samples within that task and normalizing the total to a 100-point scale. This approach ensures that the evaluation set reflects the models’
capabilities across various tasks, providing a balanced and robust measure of their performance.

\begin{table}[h]
\caption{\label{tab:eval_chat} Evaluation of Bloomz-Chat}
\begin{center}
\def\arraystretch{1.3}
\begin{tabular}{c c c c}
\hline
\textbf{Tasks} & \textbf{Samples} & \textbf{Bloomz} & \textbf{Bloomz-Chat} \\
\hline
Generic  & 10  & 49.82 & 91.25 \\
\hline
Knowledge  & 10  & 44.11 & 87.5 \\
\hline    
Roleplay  & 10  & 44.29 & 65.54  \\
\hline 
Common-Sense  & 10  & 50 & 85.07 \\
\hline 
Fermi  & 10  & 38.75 & 75.71 \\
\hline 
Counterfactual  & 10  & 47.14  & 82.5 \\
\hline 
Coding  & 7  & 5 & 31.5 \\
\hline 
Math  & 3  & 5 & 8.5  \\
\hline 
Writing  & 10  & 70 & 76.25  \\
%\hline 
\hline 
\textbf{Total}  & \textbf{80}  & \textbf{44.26} & \textbf{73.55}  \\
\hline 
\end{tabular}
\end{center}
\end{table}

\begin{table}[h]
\caption{\label{tab:eval_chat} Evaluation of GPTJ-Chat}
\begin{center}
\def\arraystretch{1.3}
\begin{tabular}{c c c c}
\hline
\textbf{Tasks} & \textbf{Samples} & \textbf{GPTJ} & \textbf{GPTJ-Chat} \\
\hline
Generic  & 10  & 31.89 & 61.11 \\
\hline
Knowledge  & 10  & 33.22 & 62.22 \\
\hline    
Roleplay  & 10  & 23 & 33.11  \\
\hline 
Common-Sense  & 10  & 25.11 & 54.22 \\
\hline 
Fermi  & 10  & 15.22 & 24.11 \\
\hline 
Counterfactual  & 10  & 15.72  & 41.11 \\
\hline 
Coding  & 7  & 5 & 5 \\
\hline 
Math  & 3  & 5 & 5  \\
\hline 
Writing  & 10  & 17.33 & 38  \\
%\hline 
\hline 
\textbf{Total}  & \textbf{80}  & \textbf{20.8} & \textbf{39.9}  \\
\hline 
\end{tabular}
\end{center}
\end{table}

The performance of Bloomz-Chat model is better than GPTJ-Chat and demonstrates acceptable performance compared to ChatGPT. For Generic, Knowledge, Common Sense tasks, the Bloomz-Chat do a good job and has a similar performance of ChatGPT. In the domains of Roleplay, Fermi, Writing, the performance is still acceptable with the score range from 65 to 76. However, in some complex tasks like Coding and Math, the performance is very poor with the score of Coding is 31.5 and Math is 8.5. The performance of GPTJ-Chat and original GPTJ are quite poor for all tasks.

For medical domain, we randomly pick 100 examples from \href{icliniq.com}{iCliniq} dataset and use the same evaluation approach.
%\begin{table}[h]
%\caption{\label{tab:eval_chat} Evaluation Tasks}
%\begin{center}
%\def\arraystretch{1.3}
%\begin{tabular}{c c c}
%\hline
%\textbf{Dataset} & \textbf{Samples} & \textbf{Scores by GPT-4} \\
%\hline
%icliniq  & 100  & 87.44  \\
%\hline
%\end{tabular}
%\end{center}
%\end{table}
The Bloomz-Doctor exhibits a noticeable performance with a score of 73.92. GPTJ-Doctor is acceptable with the score of 59.85.

In summary, the experimental results demonstrate that both Bloomz-Chat and Bloomz-Doctor models exhibit potential abilities across all tasks. This underscores the effectiveness of our approach in enhancing the Vietnamese understanding and generation capabilities of the Bloomz model. Some examples of our models display in {Table \ref{tab:result_sample}}

% \begin{table}[h]
% \caption{\label{tab:example_results} Some examples of Bloomz-Chat and Bloomz-Doctor}
% \begin{center}
% \def\arraystretch{1}
% \begin{tabular}{ | c | c| } 
%     \hline
% \textbf{Prompt} & \textbf{Response} \\
%   \hline
%   Dịch sang tiếng Việt: Je t’aime
%   & Tôi yêu bạn \\ 
%   \hline
%   Ăn gì cho đẹp da
%   & Một chế độ ăn uống lành mạnh có thể giúp bạn trông và cảm thấy tốt hơn... \\ 
% \end{tabular}
% \end{center}
% \end{table}

\begin{table*}
\caption{\label{tab:result_sample} Some examples of Bloomz-Chat and Bloomz-Doctor}
\def\arraystretch{1.5}
\begin{adjustbox}{width=\textwidth}
\begin{tabular}{p{0.3\linewidth} | p{0.65\linewidth}}
\hline
\textbf{Prompt} & \textbf{Response} \\
\hline
Dịch sang tiếng Việt: J'aime la recherche scientifique.  & Tôi yêu nghiên cứu khoa học \\
\hline
Ăn gì cho đẹp da  & 
Một chế độ ăn uống lành mạnh có thể giúp bạn trông và cảm thấy tốt hơn. Nó bao gồm trái cây, rau quả, protein nạc (như gà hoặc cá), ngũ cốc nguyên hạt và các loại thực phẩm chứa chất chống oxy hóa khác nhau như cà rốt và việt quất. Điều quan trọng là phải tránh đồ ngọt và quá nhiều chất béo bão hòa cũng được tìm thấy trong thịt đỏ và chiên\\
\hline
Ý nghĩa của cuộc sống là gì? & 
Là một mô hình ngôn ngữ AI, tôi không có ý kiến cá nhân hoặc niềm tin cá nhân. Tuy nhiên, theo các học thuyết tâm lý và thần kinh khoa học hiện đại, mục đích cơ bản của sự tồn tại được liên kết với việc tìm kiếm hạnh phúc và thỏa mãn trong những trải nghiệm này \\
\hline
Thưa bác sĩ, tôi bị sưng môi và hàm, đau nướu, sốt và loét miệng. Bạn có biết điều gì có thể gây ra tất cả các triệu chứng này không? &
Xin chào! Các nguyên nhân phổ biến nhất cho các khiếu nại mà bạn đề cập ở đây bao gồm nhiễm trùng răng hoặc nha chu (viêm nướu), viêm xoang mũi do virus hoặc vi khuẩn. Nguyên nhân khác như dị ứng với thực phẩm, thuốc lá cũng nên được loại trừ trước khi chẩn đoán chắc chắn về tình trạng hiện tại. Nếu đó là một cơn sốt virus thì sẽ mất khoảng 2 tuần để nó giảm dần hoàn toàn....\\
\hline
Thưa bác sĩ, tôi thấy mệt mỏi, khó thở, đau ngực và nhịp tim không đều. Tôi đã thực hiện một số nghiên cứu và có vẻ như tôi có thể bị viêm cơ tim. Tôi nên làm những xét nghiệm nào để xác nhận nó? & 
Xin chào! Tôi muốn biết lịch sử lâm sàng chi tiết hơn về tình trạng này (bao gồm cả việc điều trị trước đó).Tôi nghĩ rằng bạn đang nói về sự lo lắng hoặc căng thẳng trong quá trình tập luyện.Bạn cần phải kiểm tra mức độ cholesterol cao và huyết áp cũng như ECG.Điều quan trọng nhất là loại bỏ nguyên nhân gây ra vấn đề của bạn - do thiếu máu cục bộ hay tăng huyết áp? Hy vọng thông tin này hữu ích cho bạn.Cảm ơn vì đã liên hệ với chúng tôi.Chúc mọi sức khỏe tốt đẹp!Trân trọng, \\
\hline
\end{tabular}
\end{adjustbox}
\end{table*}

\section{Conclusion}

In this paper, we propose an efficient adaption method for training instruction-following language models. We also provided instruction datasets for the general and medical domains. With only 4.2M parameters and six hours of training, our approach effectively fine-tunes Bloomz and exhibits potential abilities in chatbots and medical applications. Our approach can easily be applied to other tasks and domains. However, our models still have their limits. The response from the models is sometimes not relevant to the instruction. Sometimes it responds with the wrong answer. This can be the limit of training text corpus in our base model - Bloomz - which has only 3\% Vietnamese. We believe if we can expand the pre-trained LLM with more Vietnamese data, the model performance will increase. And that will be our next work in the near future.

% \vspace{2mm}
\section*{Acknowledgment}
This research is funded by University of Information Technology-Vietnam National University HoChiMinh City under grant number D1-2023-38.

\balance

\bibliographystyle{plain}
\bibliography{ref} 

\vspace{12pt}

\end{document}